\documentclass[a4paper]{article}

\usepackage{INTERSPEECH2018}

\usepackage{algorithm}
\usepackage{algorithmicx}
\usepackage[noend]{algpseudocode}
\usepackage{xcolor}
\usepackage[font=footnotesize,labelfont=bf]{caption}
\usepackage{booktabs}
\usepackage[pagebackref=true,breaklinks=true,letterpaper=true,colorlinks,bookmarks=false]{hyperref}

\title{ Deep Lip Reading: a comparison of models and an online application }
\name{Triantafyllos Afouras, Joon Son Chung, Andrew Zisserman}
\address{
  Visual Geometry Group, Department of Engineering Science,\\
  University of Oxford, UK}
\email{\{afourast,joon,az\}@robots.ox.ac.uk}

\def\newpara{\vspace{2pt}}
\def\psubsec{\vspace{-5pt}}
\def\rsubsec{\vspace{-3pt}}

\begin{document}

\maketitle
\begin{abstract}

The goal of this paper is to develop state-of-the-art models for lip reading --
visual speech recognition.  We develop three
architectures and compare their accuracy and training times: (i) a
recurrent model using LSTMs; (ii) a fully convolutional model; and
(iii) the recently proposed transformer model. The recurrent and fully
convolutional models are trained with a Connectionist Temporal
Classification loss and use an explicit language model for decoding, the transformer is
a sequence-to-sequence model.  Our best performing model improves the
state-of-the-art word error rate on the challenging BBC-Oxford
Lip Reading Sentences 2 (LRS2) benchmark dataset by over
20 percent.

As a further contribution we investigate the fully convolutional model
when used for online (real time) lip reading of continuous speech, and show that
it achieves high performance with low latency.
\end{abstract}
\newpara\noindent\textbf{Index Terms}: lip reading, visual speech recognition


\section{Introduction}
\psubsec

In recent years, there has been a quantum leap in the performance of
visual speech recognition systems, thanks to the advances in deep
learning techniques~\cite{Krizhevsky12,Simonyan15,He15,Ioffe15} and the availability of large-scale
datasets~\cite{Chung16,Chung17}.  

In this paper, we propose three new lip reading neural network models
based on recently proposed sequence learning methods that have been
used successfully for machine translation and automatic speech recognition (ASR).  There are  two main strands in
sequence  modelling, namely using an encoder-decoder
architecture with soft-attention~\cite{Sutskever14,cho14,Bahdanau14} (`sequence to sequence'),
or using  CTC ~\cite{Graves06,graves12}. We select two models that use CTC --
a recurrent model with LSTMs, and a fully convolutional model, and from
the family of attention-based methods,  we use the recently
proposed Transformer \cite{Vaswani2017} which is the current
state-of-the-art in machine translation.

We make the following four contributions: first, we propose three
complementary new models for lip reading. For two of these, we adapt
architectures developed for other domains, namely machine translation
and ASR, and repurpose them for lip reading for the first time. 
Second, we compare the strengths and weaknesses of these architectures
in terms of performance accuracy, training time, generalization at
test time, and ease of use; third, we achieve a new state-of-the-art
on the public BBC-Oxford
Lip Reading Sentences 2 (LRS2) benchmark dataset; finally, we consider
modifications that enable on-line lip reading,  so that transcriptions are
available immediately, and not restricted to utterance-in, utterance-out.

On-line lip reading  opens up a host of new applications, such as real-time
speech captioning in noisy environments.

  \begin{figure*}[t] 
	\centering
	\vspace{-20pt}
		\includegraphics[width=\linewidth]{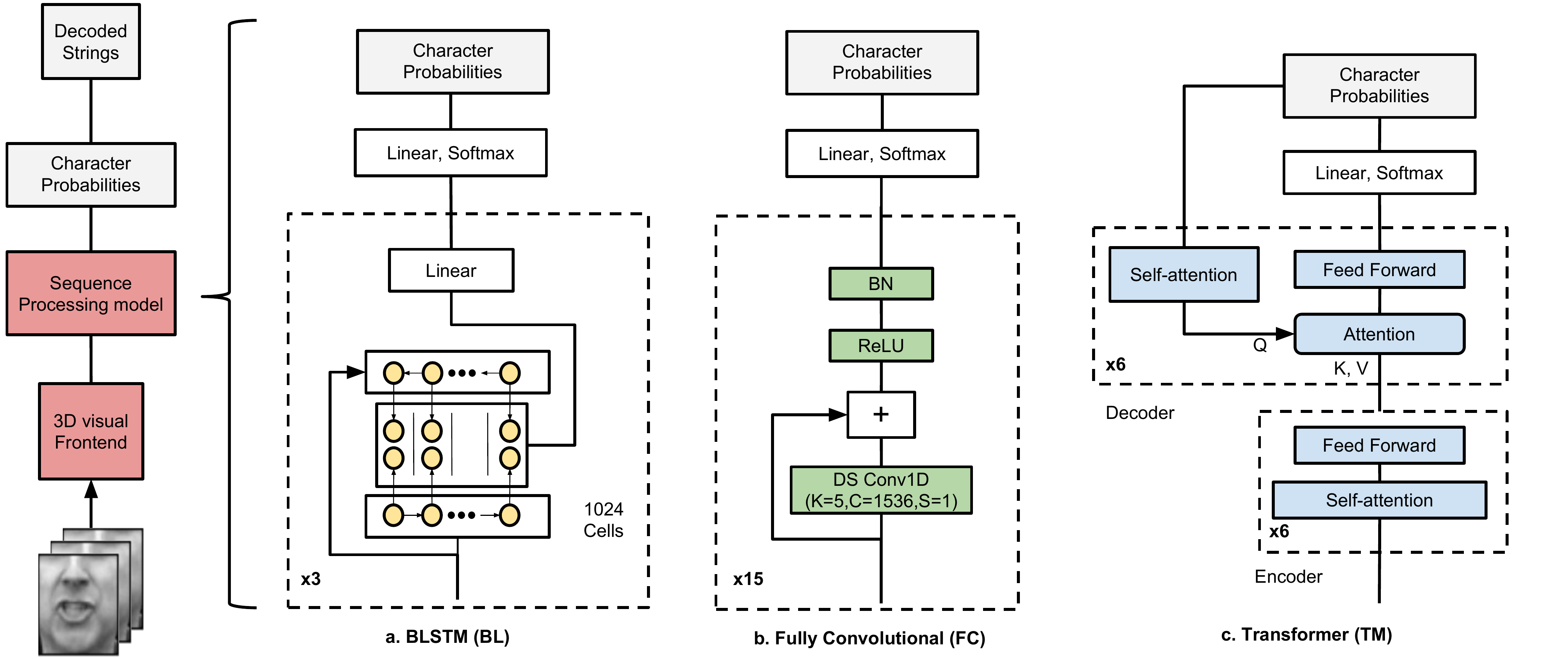}
		\vspace{-8pt}
                \caption{
                Lip reading models.
                  The image sequence is first processed by a spatio-temporal ResNet that is common
                to all models. The visual features are then processed by one of three architectures.
      \textbf{(a) BL:} The recurrent model consists of a stack of Bidirectional LSTM layers;
      \textbf{(b) FC:} The fully convolutional model is a deep  network formed of
      depth-separable convolutions.  
       \textbf{(c) TM}: a Transformer model. K, V and Q denote the Key, Value and Query tensors for the multi-head attention. 
    }
	\label{fig:pipeline}
	\vspace{-15pt}
\end{figure*}

\rsubsec
\subsection{Related works}
\psubsec
Research on lip reading has a long history, and has received an increasing amount of attention in recent years.
Large scale  datasets for lip reading are now available such as the Lip Reading in the Wild (LRW)~\cite{Chung16,Chung18} and LRS2~\cite{Chung17}.

For character-level recognition of visual sequences, the prior work can be divided into two strands.
The first strand uses CTC, where the model
predicts frame-wise labels and then looks for the optimal alignment
between the frame-wise predictions and the output sequence. 
An example based on this approach is LipNet~\cite{Assael16}, which
uses a spatio-temporal front-end, with 3D and 2D convolutions for generating the features,
followed by two  layers of BLSTM.

The second strand  is sequence-to-sequence models that first read
the input sequence before predicting the output sentence.
An example of this is the LSTM based encoder-decoder architecture with attention
 of~\cite{Chung17}, where the model can also  combine the audio and 
visual input streams. 
This work is extended in~\cite{Chung17a}, where a wider variety of poses is added to the dataset
and multi-view models are trained.

A deeper architecture than LipNet \cite{Assael16} is used by~\cite{Stafylakis17}, who propose
a residual network with 3D convolutions to extract more  powerful
representations. The network is trained with a cross-entropy loss
to recognise words from the LRW dataset.
Here, the standard ResNet architecture \cite{He15} is modified to process 3D image
sequences by changing the first convolutional and pooling blocks from 2D to 3D.
An extended version of this architecture is used for jointly modeling audio and video by~\cite{Petridis18}.

While both encoder-decoder and CTC based approaches initially relied on recurrent networks,
recently there has been a shift towards purely convolutional models \cite{Bai18}. 
For machine translation, \cite{Gehring17a} replace the encoder and \cite{Gehring17b} the whole pipeline with a fully-convolutional model.
Encoder-decoder architectures based on dilated convolutions have been
also used for translation \cite{Kalchbrenner16} and speech synthesis \cite{Oord16b}, while
\cite{Kaiser17} suggests  using depth-separable convolutions \cite{chollet2017xception} instead.
Fully convolutional networks have been recently proposed for ASR  with CTC \cite{Wang17, Zhang} or a simplified
variant \cite{Collobert16, Liptchinsky17, Zeghidour17}.

For online sequence-to-sequence prediction,  \cite{Raffel2017} uses  attention but constrains  it to be
monotonic, which allows the alignment  to be computed  online, while
 \cite{Luo17} replaces  soft with hard attention, which is trained with a policy
gradient method and does not require the whole input sequence to be available in order to start decoding.
For training online models with CTC, \cite{Kim17} use a teacher-student approach, where an offline BLSTM based model transfers
its knowledge to a unidirectional LSTM student,
while \cite{Hwang2017} use unidirectional RNNs and an expectation-maximization algorithm
dealing with long sequence lengths.
Alternatively, \cite{Jaitly16} propose a method trained with dynamic programming that conditions on the
partially observed input and allows the model to produce output online.

\section{Architectures}
\psubsec
Given a silent video of a talking face, our task is to predict the sentences being spoken. 
In this section, we propose three deep neural network models for it.
In each case the model consists of two modules (or sub-networks): a spatio-temporal visual front-end that inputs a sequence
of images of loosely cropped lip regions, and outputs one feature vector per frame; and a sequence processing
module that inputs the sequence of per-frame feature vectors and outputs a sentence character by character.
The visual front-end is common across the three models, they only differ in the sequence transcription.
We briefly describe each of these modules in the following, and illustrate them in Figure~\ref{fig:pipeline}.

\rsubsec
\subsection{Vision Module (VM)} 
\psubsec
The spatio-temporal visual front-end is based on \cite{Stafylakis17}. 
The network applies a spatio-temporal (3D) convolution on the input image sequence, 
with a filter width of five frames, 
followed by a
2D ResNet
that gradually decreases the spatial dimensions with depth (for full detail please refer to the
supplementary material).
For an input sequence of $T \times H \times W$ frames, the output is a $T \times \frac{H}{32} \times
\frac{W}{32} \times 512$ tensor ({\em i.e.}\ the temporal resolution
is preserved) that is then average-pooled over the spatial dimensions, yielding a $512$-dimensional feature
  vector for every input video frame.  

  \rsubsec
\subsection{Bidirectional LSTM (BL)}
\psubsec
 This is the first of the three
sequence transcription modules that we compare.  It consists of three
stacked bidirectional LSTM (BLSTM) recurrent layers. The first BLSTM layer
ingests the vision feature vectors, and the final BLSTM layer emits
a character probability for every input frame. The BLSTM have 1024
cells each.  The implementation of the BL network is similar to the
one used by LipNet~\cite{Assael16}.
The network is trained with CTC. The output alphabet is therefore augmented with the CTC blank
character,  and the decoding is performed with a beam search that incorporates prior information from an
external language model~\cite{Graves2014, Maas15}.

\rsubsec
\subsection{Fully Convolutional (FC) }
\psubsec
The network  consists of a number of  temporal convolutional layers.
We use depth-wise separable convolution layers~\cite{chollet2017xception},  that consist of a separate convolution
along the time dimension for every channel, followed by a projection along the channel dimensions
(a position-wise convolution with filter width 1). 
After each convolution we add a shortcut connection, followed by Batch Normalization, and ReLU.
The FC  network is also trained with a CTC loss, with sequences decoded
by using a beam search that incorporates the external language model (above).
We consider two variants: one with 10 convolutional layers (FC-10),  and a deeper one with 
15  convolutional layers (FC-15).

\begin{table*}[] 
\begin{center}
\vspace{-15pt}
\begin{tabular}{ llcccccccr } 
 \toprule
 \textbf{Net} &
 \textbf{Method} &
 \textbf{\# p}  &
 \textbf{CER Greedy} &
 \textbf{CER T2} &
 \textbf{WER Greedy} &
 \textbf{WER T1} &
 \textbf{WER T2} &
 \textbf{t/b (s)} &
 \textbf{time} \\ 

 B & MV-WAS \cite{Chung17a}    & -      & -         & -          & -        & 70.4\%   & -                & -    & -    \\      
 BL & BLSTM  + CTC             & 67M    & 40.6\%    & 38.0\%     & 76.5\%   & 62.9\%   & 62.2\%           & 0.76 & 4.5d \\ 
 FC-10 & FC$\times$10 + CTC    & 24M    & 37.1\%    & 35.0\%     & 69.1\%   & 58.2\%   & 57.1\%           & 0.23 & 2.4d \\ 
 FC-15 & FC$\times$15 + CTC    & 35M    & 35.3\%    & 33.9\%     & 64.8\%   & 56.3\%   & 55.0\%           & 0.34 & 3.4d \\     
 TM & Transformer              & 40M    & 38.6\%    & 34.0\%     & 58.0\%   & 51.2\%   &\textbf{50.0}\%   & 0.41 & 13d  \\ 
 \bottomrule
\end{tabular}
\normalsize
\vspace{-8pt}
\end{center}
\caption{
Character error rates (CER) and word error rates (WER) on the LRS2 dataset (lower is better). 
In the case of T1, we use
a LM trained on the corpus explicitly to decode the CTC models, whereas the TM model learns the corpus
implicitly during training.
For T2, the external LM is explicitly integrated at inference time for all models.
Greedy denotes decoding without beam search.
  \#p denotes the total number of parameters of the model (excluding the
  visual front-end), t/b the processing time for a single batch of 100 samples of 60 frames, and
  time the total time for completing the training curriculum on a single GPU (d=days).
  The time to train the visual front-end (2 weeks) is excluded from the statistics.
}
\label{tab:results}
\vspace{-20pt}
\end{table*}

\rsubsec
  \subsection{Transformer model (TM)}
  \psubsec
  The Transformer~\cite{Vaswani2017} model has an
  encoder-decoder structure with multi-head attention layers used as building blocks.
  The encoder is a stack of self-attention layers, where the input tensor serves as the attention queries,
  keys and values at the same time.
  Every decoder layer attends on the embeddings produced by the encoder using common soft-attention: 
  the encoder outputs are the attention keys and values and the previous decoding layer
  outputs are the queries.
  The information about the sequence order of the encoder and decoder
  inputs is fed to the model via fixed positional embeddings in the form of sinusoid functions.
  The decoder produces character probabilities which are directly matched to the ground truth labels
  and trained with a cross-entropy loss. 
  We use the base model \cite{Vaswani2017} as is, with $6$ encoder and 6 decoder layers,
  model size $512$, $8$ attention heads and dropout with $p=0.1$.
  The TM does not require an explicit language model for decoding, since it learns an implicit one
  during training on the visual sequences. However, integrating an external language model in the
  decoding process has been shown to be beneficial \cite{Kannan17}.

\rsubsec
  \subsection{External Language Model (LM) }
  \psubsec
   During inference we use a character-level language model,
   which is a recurrent network with 4 unidirectional layers of 1024 LSTM cells each.
   The LM is trained to predict one character at a time.
   Decoding is performed with a left-to-right beam search where the LM log-probabilities are combined 
   with the model's outputs via shallow fusion \cite{Kannan17}. This is common for all models,
   however the beam search is slightly more
   complicated in the CTC case.  For more details refer to the appendix.

\section{Experiments \& Results}
\label{sec:exp}
\psubsec

\subsection{Datasets and evaluation measures}
\psubsec
For training and evaluation, we use the Lip Reading in the Wild (LRW) and the
Lip Reading Sentences 2 (LRS2)
datasets.
LRW consists of approximately 489K samples,
each containing the utterance of a single word out of a vocabulary of 500.
The videos have a fixed length of 29 frames, the target word occurring in the middle of the clip and surrounded by co-articulation. All of the videos are either frontal or near-frontal.
The LRS2 dataset contains sentences of up to 100 characters
from BBC videos, with a range of viewpoints from frontal to profile.
The dataset is extremely challenging due to the variety in viewpoint, lighting conditions, genres
and the number of speakers. The training data contains over 2M word instances and
a vocabulary of over 40K.

We also make use of the MV-LRS dataset used in \cite{Chung17}, 
from which we extract individual words to obtain additional word-level pre-training data.
This auxiliary word-level set will be referred to as MV-LRS(w).
Both MV-LRS and LRS2 have ``pre-train'' sets that contain sentence excerpts which may be
shorter than the full sentences included in the train sets and are annotated with the alignment boundaries of every word.

The statistics on these datasets are summarised in Table~\ref{tab:datasets}.


\newpara\noindent\textbf{Datasets for training external language models}.
We use two different text corpora to train the language models.
The first, $T1$, only contains the transcriptions of the LRS2
pre-train and main train data ($2M$ words), and therefore the same information that is provided with teacher forcing via the decoder inputs to the
{\bf TM} model during training.
The second set, $T2$, of $26M$ words, contains the full subtitles of all the videos
from which the LRS2 training set is generated ({\em i.e.}\ $T1$ is a subset of $T2$).

\newpara\noindent\textbf{Evaluation measures.}
We evaluate the models on the LRS2 test set that consists of 1,243 utterances. 
We report Character Error Rates (CER) and Word Error Rates (WER)
on the LRS2 test set, along with the number of parameters, the computation time for a single mini-batch
and the total training time for each model.  
The error rates are defined as the normalized edit distance between the ground truth and predicted sentences. 

\rsubsec
\subsection{Training protocol}
\label{subsec:training}
\psubsec
The training proceeds in three stages: first, the visual front-end module is trained; 
second, visual features are generated for all the training data using the vision module; 
third, the sequence processing module is trained.

\newpara\noindent\textbf{Pre-training visual features}.
For the first stage, we pre-train the visual front-end on the word-level datasets (LRW and MV-LRS(w)) following
\cite{Stafylakis17}, where a 2 layer temporal convolution network is used to classify every
talking head with a word label.
The input video frames are converted to greyscale, 
scaled and centrally cropped.
We also perform data augmentation in the form of horizontal flipping, removal of random frames~\cite{Assael16,Stafylakis17}, and random shifts of up to $\pm5$ pixels in the spatial dimension and 
of $\pm2$ frames in the temporal dimension.

\newpara\noindent\textbf{Curriculum learning}.
After pre-training the visual module, we proceed with training the sequence
processing networks.
We first pass all the videos through the pre-trained front-end to obtain the visual features. 
We then train the sequence models
directly on the features, using a strategy similar to \cite{Chung17},
that starts with utterances of 2 words then of 2 and 3 words then \{2, 3, 4\} etc. 
Since the position of every word in the input video is known, we can choose any continuous sentence
excerpt contained in the dataset,
calculate the corresponding indices in the visual features sequence and load the features
extracted from the video frames containing the utterance. 
This approach helps to accelerate the training procedure.
We first train the network on the MV-LRS and 
 the ``pre-train'' part of the LRS2 dataset,
and finally fine-tune on the ``train'' set of LRS2.
We deal with the difference in utterance lengths by zero-padding them to a maximum sequence length,
which we gradually increase along with the maximum number of words used at every step of the curriculum.

\newpara\noindent\textbf{Training details}.
The {\bf TM} is trained using teacher forcing -- we supply the ground truth
of the previous decoding step as the input to the decoder, while
during inference we feed back the decoder prediction. 
The network is trained with dropout~\cite{srivastava2014dropout} with
probability $0.3$ on the inputs and the recurrent units of the BLSTM layers.
The {\bf FC} uses dropout with probability $0.8$ after each every batch normalisation layer.
For the {\bf BL} architecture we use SGD with
a fixed momentum of 0.9 and learning rate starting at $10^{-2}$ and reducing it every time the
error plateaus, down to $10^{-4}$. For the {\bf FC} and {\bf TM} we
use the ADAM optimiser~\cite{kingma2014adam} with the default
parameters and initial learning rate $10^{-3}$, reducing it on plateau down to $10^{-4}$.
All the models are implemented in TensorFlow and trained on a single 
GeForce GTX 1080 Ti GPU with 11GB memory.

\rsubsec
\subsection{Results and Model Comparison}
\psubsec
The results are summarized in Table~\ref{tab:results}.
The best performing network is the Transformer, which achieves a WER of $50\%$
 when decoded with a language model trained on $T2$, an improvement of over $20\%$ compared to the previous 
$70.4\%$
state-of-the-art~\cite{Chung17}.

\newpara\noindent\textbf{The FC model}.
The fully convolutional model has a smaller number of parameters and trains faster than {\bf BL} and
{\bf TM}, achieving 55\% WER.
Comparing to the 10-layer architecture {\bf FC-10}, the 5 additional layers contribute a $2\%$
reduction in WER. We believe this improvement to be mostly due to the wider total receptive field which gives the model more context for every prediction. 
Using depth-separable convolutions doubles the network training speed, without negatively affecting the accuracy.
With the {\bf FC} architecture, we have fine-grained control over the amount of future and
past context by adjusting the receptive field.
We cannot constrain this in the same way when using either the {\bf BL} or {\bf TM}  models, since
for both the entire input sequence needs to be available at inference time. This enables us to
perform online decoding with {\bf FC}, as described in more detail in the next section.

\newpara\noindent\textbf{The BL model}.
We obtain worse performance with {\bf BL} compared to {\bf FC-10}, even though the recurrent model has
full context on every decoding timestep compared to the convolutional that only looks at a limited
time-window of the input. 
We suspect that this is in part due to the CTC loss having a local nature: the output labels are not
conditioned  on each other and a monotonic alignment is enforced. Therefore the capacity of the
BLSTM to learn long-term, non-linear dependences cannot be fully exploited for modelling complex grammar rules.

\newpara\noindent\textbf{Language modelling.}
For all models we get an improvement of 0.7 - 1.3 \% in WER when
decoding with T2 compared to T1. 

\newpara\noindent\textbf{Training time}.
{\bf TM} and {\bf FC-15} both take approximately the same amount of time to complete a batch. Every layer of both
models has a $O(td^2)$ complexity (for $t < d$), where $d$ is the layer's width (number of channels).
{\bf TM}'s layers have smaller width (every self-attention block has a base width of 512 channels and it
is followed by two position-wise fully connected layers with 2048 and 512, compared to 1536 for the {\bf FC}), but it is
effectively a deeper model, with $3(6+6)=24$ layers in total.
However {\bf FC-15} takes fewer iterations to train, completing the full curriculum in 3.5  days,
compared to 13 days for {\bf TM}.
We hypothesize that this is due to the Transformer model being tasked
with learning the self-attention weights,
the encoder-decoder attention, and an implicit language model.
In contrast, the {\bf FC}'s task of 
learning the character-emission probabilities given a fixed context is simpler.
The {\bf BL} naturally takes more time for processing one batch, since the computations within its layers have to be run
sequentially, in contrast to the other two models.
However it converges in fewer epochs, consequently  even though the time per iteration for {\bf BL}   is almost
double that of {\bf FC-10}, it takes only one extra day to train in total. 

\newpara\noindent\textbf{Generalization to longer sequences}.
The {\bf FC} model generalises well to longer sequences once it has been trained on sentences that are long
enough to cover its full receptive field. We start observing diminishing returns in terms of accuracy gains when
training on sequences longer than 80 frames.
We had similar findings with {\bf BL}.
We could not get the {\bf TM} model to generalize as well when evaluating on longer sequences than seen during training
and, therefore we continued the curriculum in order to cover the length up
to the longest sample in the validation set.


\begin{table}[]
\begin{center}
\scriptsize
\begin{tabular}{ cl cl } 
  \toprule
 \textbf{frame \#} & \textbf{Decoded string}  & \textbf{frame \#} & \textbf{Decoded string} \\ 
  \midrule
  02& i  &                                                        02 & o\textcolor{red}{ne} \\
 04&  he \textcolor{red}{re} &                                    07 & to \textcolor{red}{} \\
  07& on \textcolor{red}{} &                                      10 & it i\textcolor{red}{n} \\
  08& a \textcolor{red}{} &                                       11 & on i\textcolor{red}{t} \\
  09& we w\textcolor{red}{hat} &                                  12 & to o\textcolor{red}{n}  \\
  10& we we\textcolor{red}{} &                                    13 & to ho\textcolor{red}{w } \\
  11& we ha\textcolor{red}{ve}&                                   14 & at home\textcolor{red}{ }\\
  12& we do\textcolor{red}{} &                                    26 & at home \textcolor{red}{}\\
  13&  we di\textcolor{red}{d} &                                  27 & at home an\textcolor{red}{d} \\
  15& we did\textcolor{red}{} &                                   28 & home \textcolor{red}{} \\
  17& we did \textcolor{red}{} &                                  29 & home to\textcolor{red}{ } \\
  18& we did i\textcolor{red}{t} &                                32 & home to \textcolor{red}{} \\
  20& we did it \textcolor{red}{} &                               33 & home to yo\textcolor{red}{ur } \\
  21& we didn't ha\textcolor{red}{ve} &                           34 & home to you\textcolor{red}{r } \\
  22& we didn't hav\textcolor{red}{e} &                           38 & home you\textcolor{red}{ } \\
 23& we did live\textcolor{red}{}  &                              40 & home you a\textcolor{red}{re } \\
 24& we didn't have \textcolor{red}{}&                            41 & home you and\textcolor{red}{ } \\
 25& we did differ\textcolor{red}{ent} &                          45 & home to you and \textcolor{red}{} \\
 27& we did differen\textcolor{red}{t} &                          46 & home to you and had\textcolor{red}{ } \\                                        
  \midrule
 \textbf{gt}& we did a different &    \textbf{gt} & home to an animal \\
  \bottomrule
\end{tabular}                                              
\end{center}    
\vspace{-8pt} 
\normalsize                                          
\caption{Online decoding examples. Red color denotes the completions of words by the language
model. The last line contains the ground truth transcriptions of the excerpt. }   
\label{tab:online_decoding}                   
\vspace{-30pt}             
\end{table}                                                

\section{Online lip reading}
\psubsec
\label{sec:online}
In this section we describe how the {\bf FC} model can be used for online lip reading with low latency.
One advantage of using the temporal convolutions is that we can control how much future context
we want to allow the model to see. In contrast, when using bidirectional recurrent networks, or any
model with vanilla attention, the entire input sequence needs to be available at the start of the inference. 
Every temporal convolution with filter width $K$ contributes $\frac{K-1}{2}$ future frames
to the overall receptive field.  The total receptive field of a network with $L$ similar layers
is $ R = L\times \frac{K-1}{2}\times2+1$ frames,
which allows it to peek up to $r = L\times \frac{K-1}{2}$ frames into the future. 
In our setting with $K=5$, $r$ is equal to $22$ and $32$ frames for the 11 and 16 layer
models respectively (here we also take into account the contribution of the front-end's 3D convolutions).

Training with CTC is known to result in peaky distributions ~\cite{Zeyer17,chen17b,Rosenberg17}.  
In practice we find that the network emits a character with high probability  
when the frames that trigger it are under the center of its receptive field. 
In an online setting we would receive one input video frame at a time.
To obtain the same decodings as when running offline,
it is sufficient to apply the convolutions on the incoming frames with a time lag of $r$ frames:
At the decoding time step $t$ the network's receptive field is centred at frame $t-r$ and emits
a distribution $p^{ctc}_t$, peeking $r$ frames into the future.
The beam search step can be run iteratively on the probabilities $p^{ctc}_t$,
scoring them with the language model and accumulating them into the running hypotheses.
The final prediction is the same as the offline case.

However, since the network is trained on variable length inputs, it is able to handle partial sentences. 
For every decoding time-step of the loop described above, we can run additional $r$ beam search
steps as if the sentence would end at the current frame.
In this manner, we can make predictions in real time on every time step with an additional computation overhead proportional to the size of the receptive field.
Using convolutions requires only $O(r)$ new computations for the network forward pass to obtain the CTC
emission probabilities and then an extra $O(r W |A|)$ to run the Beam Search, where $|A|$ is the alphabet
size and $W$ the beam search width,  overall resulting in linear time complexity, $O(T r W |A|)$.
We summarize the procedure in Algorithm~\ref{alg:conv_online} in the appendix.

Finally, on every decoding time step we can predict further into the future by querying the language model. 
We show examples of online decoding in Table~\ref{tab:online_decoding}, where the endings of incomplete words of
the current beam state are filled in by the language model.

\section{Conclusion}
\psubsec
We have proposed and compared three new neural network architectures for lip reading, 
and exceeded the previous state-of-the-art by a large margin. The networks will be publicly
released. We have also carried out a preliminary investigation of on-line lip reading and 
proposed a decoding algorithm for this.
Future work could include varying the activations (e.g.\  Maxout or PReLU as in \cite{Zhang}).
Another strand to investigate is whether outputting phonemes and byte-pairs
rather than characters, as is now standard for ASR, would lead to a boost in performance.

\newpara\noindent\textbf{Acknowledgements.}
Funding for this research is provided by the UK EPSRC
CDT in Autonomous Intelligent Machines and Systems, 
the Oxford-Google DeepMind Graduate Scholarship, and by the EPSRC 
Programme Grant Seebibyte EP/M013774/1. 


\bibliographystyle{IEEEtran}

\bibliography{shortstrings,vgg_local,vgg_other,mybib}

\clearpage
\appendix

\section{Appendix}

\subsection{Dataset statistics}

The statistics of the datasets used in this paper is given in Table~\ref{tab:datasets}.

\begin{table}[ht] 
\begin{center}
\begin{tabular}{ lcrrr } 
 \toprule
 \textbf{Name} & \textbf{Type} & \textbf{Vocab} & \textbf{\#Utter.} & \textbf{\#Words}  \\ 
 LRW       & word  & 500    & -  & 489K \\ 
 MV-LRS(w) $*$   & word  & 480    & -  & 1,9M \\ 
 \midrule

 MV-LRS $*$   & sent. & 30K & 430K & 5M   \\ 
 LRS2       & sent. & 41K & 142K & 2M   \\ 
 \midrule
 T1       & text  & 41K & 142K & 2M   \\ 
 T2 $*$       & text  & 60K & 8M & 26M   \\ 
 \bottomrule
\end{tabular}
\normalsize
\end{center}
\caption{ Description of the datasets used for training and testing. 
  We formed MV-LRS(w) by isolating individual word excerpts of the 480 most frequent words, all of which have a count of at least
  1000 samples.
  The statistics for the MV-LRS and the LRS2 datasets include the noisy ``pre-train'' sets in addition to the main dataset.
  T1 consists of the transcriptions of the samples in LRS2. We form T2 by collecting
  the full transcripts of all the subtitles of the shows used in the making of LRS2. The sets marked
  with $*$ are not publicly available.\\
}
\label{tab:datasets}
\vspace{-20pt}
\end{table}

\subsection{Visual front-end architecture}

The details of the spatio-temporal front-end are given in Table~\ref{tab:resnet_arch}.

\begin{table}[!h]
\begin{center}
\scriptsize
\begin{tabular}{ lll } 
 \toprule
 \textbf{Layer Type} & \textbf{Filters} & \textbf{Output dimensions} \\ 
 \midrule
 Conv 3D & $5\times7\times7$, $64$, $/ [1, 2, 2]$  & $T\times \frac{H}{2} \times \frac{W}{2} \times 64$ \\
\addlinespace[0.5em]
 Max Pool 3D & $/ [1, 2, 2]$                       & $T\times \frac{H}{4} \times \frac{W}{4} \times 64$ \\
 \midrule
 Residual Conv 2D  & [$3\times3$, $64$] $\times2$  $/  1$  & $T\times \frac{H}{4} \times \frac{W}{4} \times 64$ \\
\addlinespace[0.5em]
 Residual Conv 2D  & [$3\times3$, $64$] $\times2$  $/  1$  & $T\times \frac{H}{4} \times \frac{W}{4} \times 64$ \\
 \midrule                                                              
 Residual Conv 2D  & [$3\times3$, $128$] $\times2$  $/ 2$  & $T\times \frac{H}{8} \times \frac{W}{8} \times 128$ \\
\addlinespace[0.5em]
 Residual Conv 2D  & [$3\times3$, $128$] $\times2$  $/ 1$  & $T\times \frac{H}{8} \times \frac{W}{8} \times 128$ \\
 \midrule                                                             
 Residual Conv 2D  & [$3\times3$, $256$] $\times2$  $/ 2$  & $T\times \frac{H}{16} \times \frac{W}{16} \times 256$ \\
\addlinespace[0.5em]
 Residual Conv 2D  & [$3\times3$, $256$] $\times2$  $/ 1$  & $T\times \frac{H}{16} \times \frac{W}{16} \times 256$ \\
 \midrule                                                             
 Residual Conv 2D  & [$3\times3$, $512$] $\times2$  $/ 2$  & $T\times \frac{H}{32} \times \frac{W}{32} \times 512$ \\
\addlinespace[0.5em]
 Residual Conv 2D  & [$3\times3$, $512$] $\times2$  $/ 1$  & $T\times \frac{H}{32} \times \frac{W}{32} \times 512$ \\
 \midrule  

\end{tabular}
\normalsize
\end{center}
\caption{Architecture details for the spatio-temporal visual front-end \cite{Stafylakis17}. The
strides for the residual 2D convolutional blocks apply to the first layer of the block only (i.e. the
total down-sampling factor in the network is 32). A short cut connection is added after every pair of
2D convolutions \cite{He15}. The 2D convolutions are applied separately on every time-frame.}
\label{tab:resnet_arch}
\vspace{-20pt}
\end{table}


\subsection{Seq2Seq decoding with external language model}
For decoding with the TM model, we use a left-to right beam search with width $W$ as in~\cite{Kannan17,Wu16},
with the hypotheses $y$ being scored as follows: 
\begin{equation}
  score(x,y) = \frac { log \ p(y|x) + \alpha \ log \ p_{LM}(y) }{ LP(y) } \nonumber
\end{equation}
where $p(y|x)$ and $p_{LM}(y)$ are the probabilities obtained from the visual and language models respectively and
LP is a length normalization factor LP(y) = $ \Big( \frac{5+|y|}{6}\Big)^\beta $ \cite{Wu16}. We did not experiment
with a coverage penalty.
The best values for the hyperparameters were determined via grid search on the validation set: for
decoding without the external language model (T1) they were set to  $W=5$, $\alpha=0.0$,
$\beta=0.6$ and for decoding with the LM (T2) to $W=15$, $\alpha=0.1$ $\beta=0.7$.

\subsection{CTC decoding algorithm with external language model}

Algorithm~\ref{alg:beam_lm} describes the CTC decoding procedure with an external language model.
It is also a beam search with width W and hyperparameters $\alpha$ and $\beta$ that control the
relative weight given to the LM and the length penalty. The beam search is similar to the one
described for seq2seq above, with some additional bookkeeping required to handle the emission of
repeated and blank characters and normalization LP(y) = $ |y|^\beta$.
We obtain the best results with $W=100$, $\alpha=0.5$, $\beta=0.1$.

\begin{algorithm}[!h]
  \caption{CTC Beam search decoding with Language Model adapted from \cite{Maas15}. 
    Notation: 
    A is the alphabet;
    $p_b(s,t)$ and $p_{nb}(s,t)$ are the probabilities of partial output transcription s resulting from paths
  ending in blank and non-blank token respectively, given the input sequence up to time $t$; $p(s,t) = p_b(s,t) + p_{nb}(s,t)$. }
   \label{alg:beam_lm}
   \hskip -2em
\begin{algorithmic} 

  \State \textbf{Parameters} CTC probabilities $p^{ctc}_{1:T}$, word dictionary, beam width $W$,
  hyperparameters $\alpha$, $\beta$

  \State initialize $\mathbf{B_{t}} \leftarrow$ \{$\varnothing$\}; 
  $\mathbf{p_b(\varnothing, 0)} \leftarrow$ 1; $\mathbf{p_{nb}(\varnothing, 0)} \leftarrow$ 0 

  \For{$t=1$ {\bfseries to} $T$} 

    \State $\mathbf{B_{t-1}} \leftarrow$ $W$ prefixes with highest $\frac{ log \ p(s,t)}{|s|^\beta}$ in $\mathbf{B}_t$
    \State $\mathbf{B_t} \leftarrow$ \{\}

    \For{prefix $s$ {\bfseries in} $\mathbf{B_{t-1}}$} 

      \State $ c^{-} \leftarrow  $ last character of $s$

      \State  $p_b(s,t) \leftarrow p^{ctc}_t(-,t) p(s,t-1) $  \Comment{adding a blank}
      \State $p_{nb}(s,t) \leftarrow p^{ctc}_t(c^-,t) p_{nb}(s,t-1)$  \Comment{repeated}

      \State add $s$ to $\mathbf{B}$

      \For{character $c$ {\bfseries in} $A$} 

        \State $ s^{+} \leftarrow  s + c$
        
        \If {$s$ ends in $c$}
          \State  $p_{c} \leftarrow p^{ctc}_t(c,t) p(c,t-1) p_{LM}(c|s)^\alpha $  
        \Else
          \State \Comment{repeated chars must have blanks in between}
          \State  $p_{c} \leftarrow p^{ctc}_t(c,t) p_{b}(c,t-1) p_{LM}(c|s)^\alpha $  
        \EndIf 

        \If {$s^{+}$ is already in $\mathbf{B_t}$ }
        \State $p_{nb}(s^+,t) \leftarrow p_{nb}(s^+,t) + p_{c}$

        \Else
          \State add $s^+$ to $\mathbf{B_t}$
          \State $p_{nb}(s,t) \leftarrow 0$ 
          \State $p_{nb}(s^+,t) \leftarrow p_{c}$
        \EndIf

      \EndFor

    \EndFor

\EndFor
        
\State \Return $ max_{s \in B_t} \frac{ log \ p(s,T)}{|s|^\beta}$ in $\mathbf{B}_T$

\end{algorithmic}
\end{algorithm}


 \newpage
\subsection{Online CTC decoding algorithm}

Algorithm~\ref{alg:conv_online} describes the online CTC decoding procedure introduced in Section~\ref{sec:online}.

\begin{algorithm}[th]
  \caption{ 
  Online CTC decoding with fully convolutional model. The algorithm runs in $O(TrW|A|)$
    time, where $T$ denotes the input sequence length, $r$ is half the length of the network's total receptive field,
    W the beam width and $|A|$ the number of characters in the alphabet.
  The BeamStep routine performs one step of the CTC Beam Search decoding outer loop shown in Algorithm \ref{alg:beam_lm}}. 
   \label{alg:conv_online}
   \hskip -2em
\begin{algorithmic} 

  \State \textbf{Parameters} Input video frames $x_{1:T}$, FC network $f_\theta$

  \State initialize $\mathbf{B_{0}} \leftarrow$ \{$\varnothing$\}; $\mathbf{L_{0}} \leftarrow$ \{$\varnothing$\}; \Comment Beam \& LM states

  \For{$t=1$ {\bfseries to} $T$}  \Comment decoding steps lag by $r$ behind real time

  \State $p^{ctc}_{t:t+r} \leftarrow f_\theta (x_{t-r:t})$  \Comment Slide network right by one step
  \State $\mathbf{B_{t}, L_{t}} \leftarrow $ \Call{BeamStep}{ $p^{ctc}_{t}, \mathbf{B_t, L_t}$ }
  \Comment $O(W \cdot |A|)$

  \State $\mathbf{\hat{B}_{t}, \hat{L}_{t}} \leftarrow$ copy $\mathbf{B_{t}, L_{t}} $   

    \For{$\tau=t+1$ {\bfseries to} $t+r$} 

      \State $\mathbf{\hat{B}_{\tau}, \hat{L}_{t}} \leftarrow $ \Call{BeamStep}{ $p^{ctc}_{\tau}, \mathbf{\hat{B}_\tau, \hat{L}_\tau}$ } 
  \Comment $O(W \cdot |A|)$

    \EndFor

    \State $D_t \leftarrow  \text{highest scoring sentence } S \in \hat{B}_{t+r} $ 

  \EndFor
        
\State \Return $ D_T$

\end{algorithmic}
\end{algorithm}

\subsection{Confusion Matrix}

\begin{figure}[!h] 
  \centering
          \includegraphics[width=1\columnwidth]{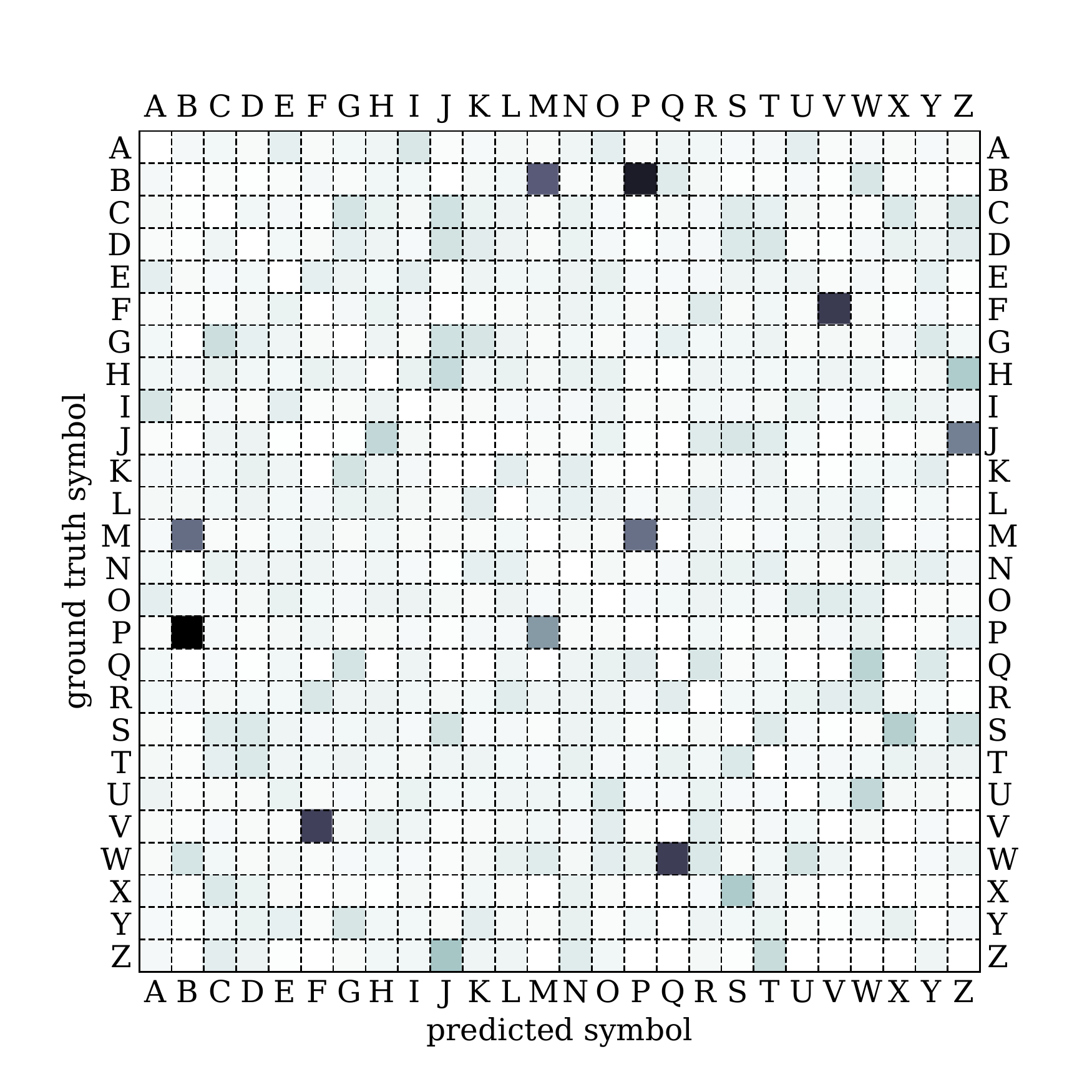}
          \caption{ Confusion matrix of CTC predictions.
        The entries of the table are the normalized substitution counts obtained
      from the minimum edit distance calculation between ground truth and sentences predicted with
      greedy CTC decoding, averaged over the
      whole dataset. 
      We observe that the network confuses characters that are difficult
  to discriminate between using the visual information alone. For example $B$ is frequently confused with $M$ and $P$,
  while $V$ is confused with $F$ and vice versa. It is interesting to note that $Q$ is often emitted instead of $W$.
 We hypothesize that this happens because these characters appear very similar visually in words like 'week / quick', 'woe / quote'. }
  \label{fig:ctc_conf}
\end{figure}

Figure~\ref{fig:ctc_conf} shows the confusion between the predictions of the FC-15 model obtained
with greedy decoding of the CTC posteriors.

\subsection{Decoding examples}

Table~\ref{tab:online_decoding_full} shows further examples of online decoding outputs.

\begin{table*}[!b]
\begin{center}
\small
\begin{tabular}{ cl l } 
  \toprule
  \textbf{frame \#} & \textbf{Decoded string}  & \textbf{Decoded string starting from the middle (frame \#55)} \\ 
  \midrule
 002 & o\textcolor{red}{ne}                                 & \\                            
 007 & to \textcolor{red}{}                                 &\\
 010 & it i\textcolor{red}{n }                              &\\
 011 & on i\textcolor{red}{t }                              &\\
 012 & to o\textcolor{red}{n }                              &\\
 013 & to ho\textcolor{red}{w }                             &\\
 014 & at home\textcolor{red}{ }                            &\\
 026 & at home \textcolor{red}{}                            &\\
 027 & at home an\textcolor{red}{d}                         &\\
 028 & home \textcolor{red}{}                               &\\
 029 & home to\textcolor{red}{ }                            &\\
 032 & home to \textcolor{red}{}                            &\\
 033 & home to yo\textcolor{red}{ur }                       &\\
 038 & home you\textcolor{red}{ }                           &\\
 040 & home you a\textcolor{red}{re}                        &\\
 041 & home you and\textcolor{red}{ }                       &\\
 045 & home to you and \textcolor{red}{}                    &\\
 046 & home to you and had\textcolor{red}{}                 &\\
 047 & home you and had\textcolor{red}{ }                   &\\
 048 & home you and ad\textcolor{red}{am }                  &\\
 051 & home you and anima\textcolor{red}{ls }               &\\
 054 & home to an animal\textcolor{red}{ }                  &\\

056 & home you and animal\textcolor{red}{s }                                         & i\textcolor{red}{}                                       \\ 
058 &                                                                                &  i\textcolor{red}{n }                                    \\                          
059 &                                                                                & it i\textcolor{red}{n }                                  \\ 
060 & home to an animal \textcolor{red}{      }                                      & i a\textcolor{red}{nd }                                  \\ 
061 & home to an animal and\textcolor{red}{    }                                     & in \textcolor{red}{}                                     \\ 
062 &                                                                                & in th\textcolor{red}{e }                                 \\ 
063 & home to an animal th\textcolor{red}{at     }                                   & then the\textcolor{red}{ }                               \\ 
064 &                                                                                & that i\textcolor{red}{s }                                \\ 
066 & home to an animal that \textcolor{red}{     }                                  & that he\textcolor{red}{re }                              \\ 
067 & home to an animal that i\textcolor{red}{s    }                                 &                                                          \\ 
068 &                                                                                & that i\textcolor{red}{s }                                \\ 
070 &                                                                                & that it's\textcolor{red}{}                               \\ 
070 &                                                                                & that is\textcolor{red}{ }                                \\ 
074 & home to an animal that is \textcolor{red}{    }                                & that it's\textcolor{red}{}                               \\ 
075 & home to an animal that is ri\textcolor{red}{ght}                               & that it's rig\textcolor{red}{ht }                        \\ 
076 & home to an animal that it's rig\textcolor{red}{ht}                             &                                                          \\ 
078 & home to an animal that is right\textcolor{red}{   }                            &                                                          \\ 
081 & home to an animal that is right in\textcolor{red}{ }                           & that is right in\textcolor{red}{ }                       \\ 
082 &                                                                                & that it's right in\textcolor{red}{}                      \\ 
083 & home to an animal that is right in t\textcolor{red}{he}                        & that it's right in t\textcolor{red}{he }                 \\ 
087 & home to an animal that is right in the t\textcolor{red}{raining}               & that it's right in the t\textcolor{red}{own }            \\ 
089 & home to an animal that is right in the to\textcolor{red}{wn     }              &                                                          \\ 
090 & home to an animal that it's right in the top\textcolor{red}{     }             & that it's right in the top\textcolor{red}{   }           \\ 
091 & home to an animal that it's right in the top \textcolor{red}{     }            & that it's right in the top \textcolor{red}{   }          \\ 
092 & home to an animal that it's right in the top o\textcolor{red}{f}              & that it's right in the top o\textcolor{red}{f  }          \\
094 & home to an animal that it's right in the top of \textcolor{red}{}             & that it's right in the top of \textcolor{red}{  }         \\
097 & home to an animal that it's right in the top of th\textcolor{red}{e}          & that it's right in the top of th\textcolor{red}{e}        \\
098 & home to an animal that it's right in the top of the r\textcolor{red}{oom }    & that it's right in the top of the fr\textcolor{red}{ont } \\
099 &                                                                                & that it's right in the top of the ro\textcolor{red}{om } \\ 
100 & home to an animal that it's right in the top of the foo\textcolor{red}{d}      & that it's right in the top of the foo\textcolor{red}{t } \\ 
101 &                                                                                & that it's right in the top of the food\textcolor{red}{ } \\ 
102 & home to an animal that it's right in the top of the foot\textcolor{red}{}      & that it's right in the top of the foot\textcolor{red}{ } \\
103 & home to an animal that it's right in the top of the futur\textcolor{red}{e}   & that it's right in the top of the futur\textcolor{red}{e} \\    
  \midrule
\textbf{\# changes/frame } & 0.4  & 0.5 \\ 
  \bottomrule

\end{tabular}                                              
\normalsize
 \end{center}                                               
 \caption{ Example of sequential online decoding starting the beginning (\textbf{left}) and from the middle of the utterance (\textbf{right}).
   The ground truth transcription is ``home to an animal that is right at the top of the food chain''.
   Red color denotes the completions of words by the language model. It can be seen that after some initial frames where the model does not have enough context
   to make a confident prediction, it starts predicting correctly.
 }   
 \label{tab:online_decoding_full}                                
 \end{table*}

\end{document}